\documentclass[conference]{IEEEtran}
\IEEEoverridecommandlockouts
\usepackage{subfigure}
\usepackage{float}
\usepackage{booktabs}
\usepackage{cite}
\usepackage{amsmath,amssymb,amsfonts}
\usepackage{algorithmic}
\usepackage{graphicx}
\usepackage{textcomp}
\usepackage{xcolor}
\def\BibTeX{{\rm B\kern-.05em{\sc i\kern-.025em b}\kern-.08em
    T\kern-.1667em\lower.7ex\hbox{E}\kern-.125emX}}
\begin{document}
\title{ORB-SLAM3AB: Augmenting ORB-SLAM3 to Counteract Bumps with Optical Flow Inter-frame Matching\\
\thanks{This work was supported by the National Science Foundation of China 62271392, Shaanxi Innovation Team 2023CXTD04. Corresponding author: Chen He(chenhe@nwu.edu.cn).\\
    \textsuperscript{1}These authors contribute equally to this paper and should be considered as co-first author.}
}

\author{\IEEEauthorblockN{1\textsuperscript{st} Yangrui Dong\textsuperscript{1}}
\IEEEauthorblockA{
\textit{Information Science}\\\textit{Northwest University} \\
Xi,an China \\
DongYangRui@stumail.nwu.edu.cn}
\and
\IEEEauthorblockN{1\textsuperscript{st}  Weisheng Gong\textsuperscript{1}}
\IEEEauthorblockA{
\textit{Information Science}\\
\textit{Northwest University} \\
Xi,an China \\
GongWeiSheng@stumail.nwu.edu.cn}
\and
\IEEEauthorblockN{3\textsuperscript{rd} Qingyong Li}
\IEEEauthorblockA{\textit{Information Science}\\
\textit{Northwest University} \\
Xi,an China \\
liqingyong@stumail.nwu.edu.cn}
\and
\hspace*{1.5cm}\IEEEauthorblockN{4\textsuperscript{th} Kaijie Su}
\IEEEauthorblockA{\hspace*{1.5cm}\textit{Information Science}\\
\hspace*{1.5cm}\textit{Northwest University} \\
\hspace*{1.5cm}Xi,an China \\
\hspace*{1.5cm}dearmoonnn@163.com}
\and
\hspace*{1.5cm}\IEEEauthorblockN{5\textsuperscript{th} Chen He*}
\IEEEauthorblockA{\hspace*{1.5cm}\textit{Information Science}\\
\hspace*{1.5cm}\textit{Northwest University} \\
\hspace*{1.5cm}Xi,an China \\
\hspace*{1.5cm}chenhe@nwu.edu.cn}
\and
\IEEEauthorblockN{6\textsuperscript{th} Z. Jane Wang}
\IEEEauthorblockA{\hspace*{0.5cm}\textit{Department of Electrical and Computer Engineerin}\\
\textit{ The University of British Columbia} \\
Vancouver, BC V6T1Z4, Canada \\
zjanew@ece.ubc.ca}
}

\vspace{-25pt}
\maketitle
\vspace{-50pt}
\begin{abstract}
This paper proposes an enhancement to the ORB-SLAM3 algorithm, tailored for applications on rugged road surfaces. Our improved algorithm adeptly combines feature point matching with optical flow methods, capitalizing on the high robustness of optical flow in complex terrains and the high precision of feature points on smooth surfaces. By refining the inter-frame matching logic of ORB-SLAM3, we have addressed the issue of frame matching loss on uneven roads. To prevent a decrease in accuracy, an adaptive matching mechanism has been incorporated, which increases the reliance on optical flow points during periods of high vibration, thereby effectively maintaining SLAM precision.
Furthermore, due to the scarcity of multi-sensor datasets suitable for environments with bumpy roads or speed bumps, we have collected LiDAR and camera data from such settings. Our enhanced algorithm, ORB-SLAM3AB, was then benchmarked against several advanced open-source SLAM algorithms that rely solely on laser or visual data. Through the analysis of Absolute Trajectory Error (ATE) and Relative Pose Error (RPE) metrics, our results demonstrate that ORB-SLAM3AB achieves superior robustness and accuracy on rugged road surfaces.

\end{abstract}

\vspace{-10pt}
\section{Introduction}
SLAM (Simultaneous Localization and Mapping) is crucial for robotics and autonomous driving, providing essential capabilities for accurate navigation and environment mapping \cite{b1,b2,b3,aulinas2008slam,dissanayake2001solution,engel2014lsd,kohlbrecher2011flexible}. SLAM systems are mainly categorized into laser-based and visual SLAM.\cite{Tourani_2022,huang2021review,singandhupe2019review,kazerouni2022survey,macario2022comprehensive,sumikura2019openvslam,strasdat2012visual,yousif2015overview,debeunne2020review,cattaneo2022lcdnet,chen2021overlapnet,park2018elastic} Laser SLAM methods, such as Livox-SLAM \cite{b24}, R3LIVE \cite{b32}, and CT-ICP \cite{b31}, are known for their stability and accuracy. In contrast, visual SLAM methods, including ORB-SLAM3 \cite{b13} and VINS-Fusion \cite{b36}, utilize texture information and techniques like feature matching and optical flow for high-precision localization and mapping.\cite{taketomi2017visual}

The origins of pure visual SLAM can be traced back to Davison's Mono SLAM\cite{davison2007monoslam} in 2007. Significant advancements were made between 2015 and 2017 when Mur-Artal and his colleagues developed ORB-SLAM\cite{mur2015orb}  and ORB-SLAM2 \cite{mur2017orb}. In 2016, J. Engel, V. Koltun and D. Cremers proposed Direct Sparse Odometry\cite{engel2016dso}. In 2021, Carlos Campos, Richard Elvira, and their colleagues introduced ORB-SLAM3, which accommodates different types of cameras and enhances pose estimation, further advancing the development of visual SLAM technology. 
\begin{figure*}[htbp]
  \vspace{-10pt}
   \centering
    \subfigure{\includegraphics[width=15cm,height=8cm]{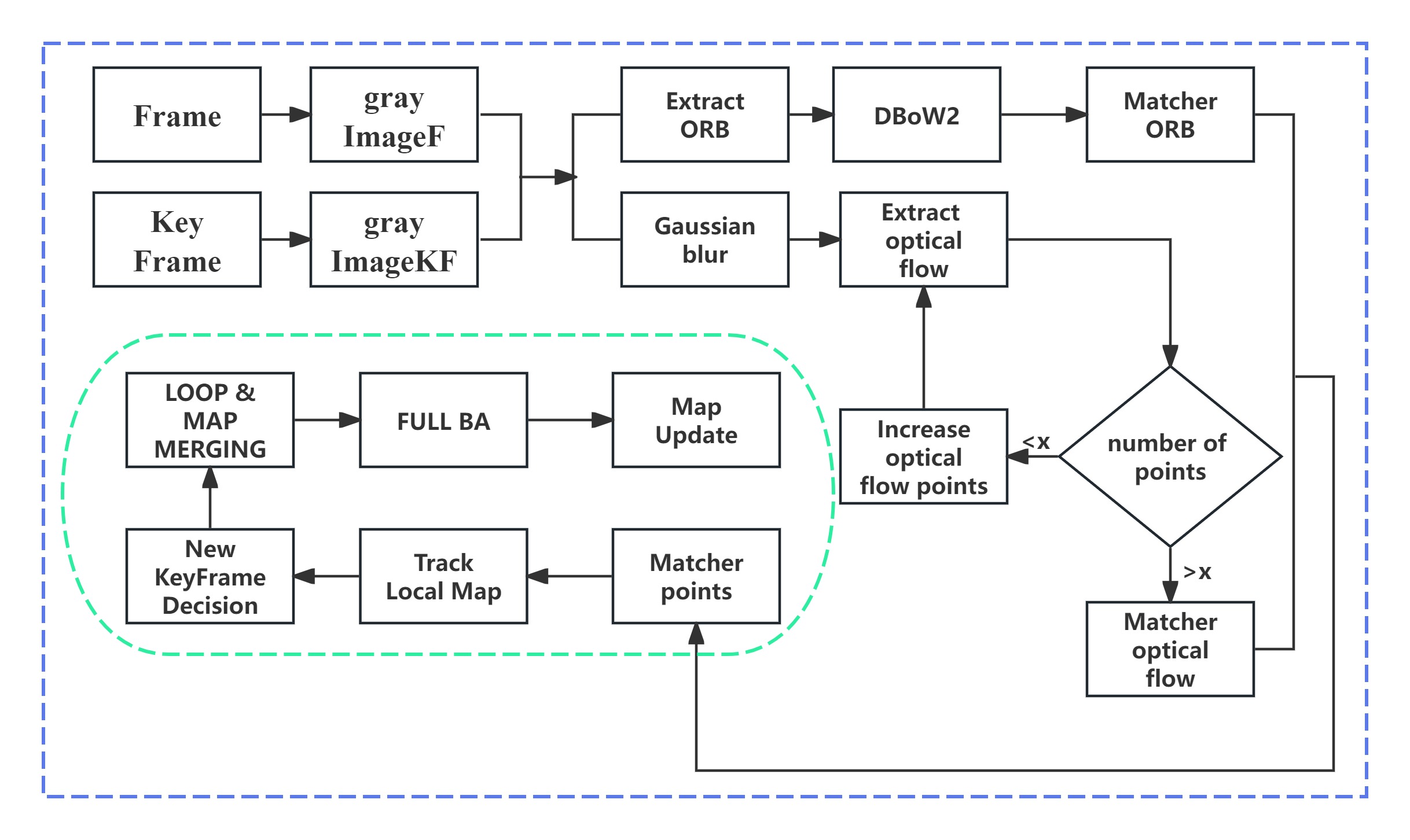}} 
  \caption{We improve visual SLAM robustness by converting frames to grayscale, then processing them with ORB extraction and optical flow. Gaussian filtering reduces noise before applying optical flow. To maintain ORB feature accuracy, we start with a minimal point selection for optical flow, increasing it only with significant camera shake. The green box represents the original ORB-SLAM3 framework.}
    \label{fig:data_distribution}
    \vspace{0.2in}
\end{figure*}
Among these algorithms, only a few, such as ORB-SLAM, support high-precision monocular camera SLAM, and most of them do not. Moreover, these monocular SLAM algorithms have neglected the research and development of robustness against bumpy ground. In 2024, Zhang Xiao and Li Shuaixin \cite{xiao2024sl} proposed SL-SLAM, replacing traditional feature point selection and matching with deep learning techniques. However, this paper has not yet been published or had its code made public.

The work most closely related to ours is the ORB-SLAM2S\cite{diao2021orb}: A Fast ORB-SLAM2 System with Sparse Optical Flow Tracking proposed by Y. Diao, R. Cen, and F. Xue in 2021. However, their method focuses on improving the running speed through optical flow without enhancing the accuracy of SLAM in bumpy scenarios. Moreover, they use optical flow for non-keyframes and feature point methods for keyframes, which is significantly different from our designed integrated approach. In contrast, our algorithm enhances the robustness and precision of SLAM during bumpy conditions and has also been compared with laser SLAM on the same route.

In 2023, Zhao Tong and He Junxiang \cite{zhao2023comprehensive} introduced a dataset focused on road surface information, along with auxiliary algorithms for intelligent driving. In 2024 \cite{zhao2024road}, Zhao Tong and Xie Yichen released the RSRD dataset, emphasizing complex road conditions with LiDAR and visual sensor data. However, datasets aimed at detecting ground unevenness pose significant challenges when used for SLAM.

This paper proposes a method that enhances the robustness and accuracy of ORB-SLAM3 under monocular conditions by integrating frame-to-frame matching based on optical flow with its existing feature-based approach.  Using an adaptive algorithm to manage optical flow features, our approach improves ORB-SLAM3’s performance on bumpy roads. \cite{post2003state,beauchemin1995computation,barron1992performance,horn1981determining,sun2010secrets,tareen2018comparative,chien2016use,gupta2019improved}The comparison results with various monocular SLAM, pure LiDAR SLAM, and LiDAR-IMU combined SLAM algorithms demonstrate the superior effectiveness of our proposed method.

Our main contributions are as follows:
\begin{enumerate}
    \item We enhanced ORB-SLAM3's frame-to-frame matching by integrating optical flow techniques, overcoming the limitations of feature-based matching and boosting robustness in scenarios with bumps and sharp turns. 
    \item We conducted a comprehensive comparative analysis of our improved algorithm with both laser-based and pure monocular SLAM methods. This included plotting trajectories and calculating both Absolute Trajectory Error (ATE) and Relative Pose Error (RPE), with results demonstrating that our method excels in environments with intense bumps and significantly enhances the accuracy of monocular SLAM.
    \item We collected data from low-speed robots and high-speed vehicles in bumpy and multi-speed-bump scenarios, providing a valuable resource for assessing SLAM performance under these challenging conditions. 
\end{enumerate}

\section{ORB-SLAM3AB Framework}

The ORB-SLAM3AB framework represents an advanced extension of the original ORB-SLAM3, designed to enhance the robustness and accuracy of monocular SLAM systems, particularly in challenging conditions such as bumpy roads. ORB-SLAM3AB improves the ORB-SLAM3 framework by incorporating enhanced feature point selection methods and modifying the feature point frame-to-frame matching strategy, addressing the limitations faced by visual SLAM in practical applications.For the process flow, please refer to Figure 1. 

  \begin{figure*}[htbp]
   \centering
    \includegraphics[width=0.48\textwidth]{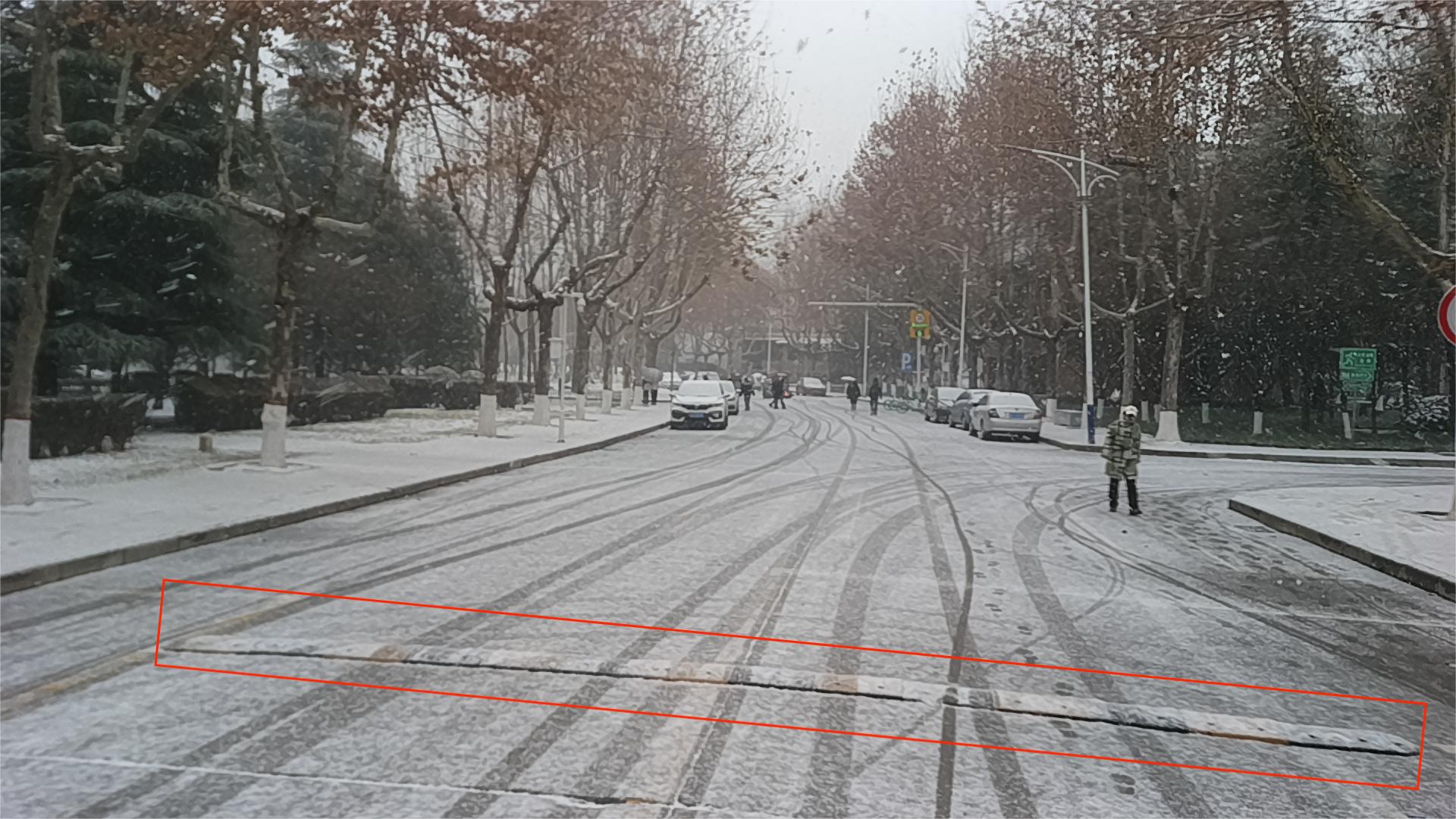}
    \includegraphics[width=0.48\textwidth]{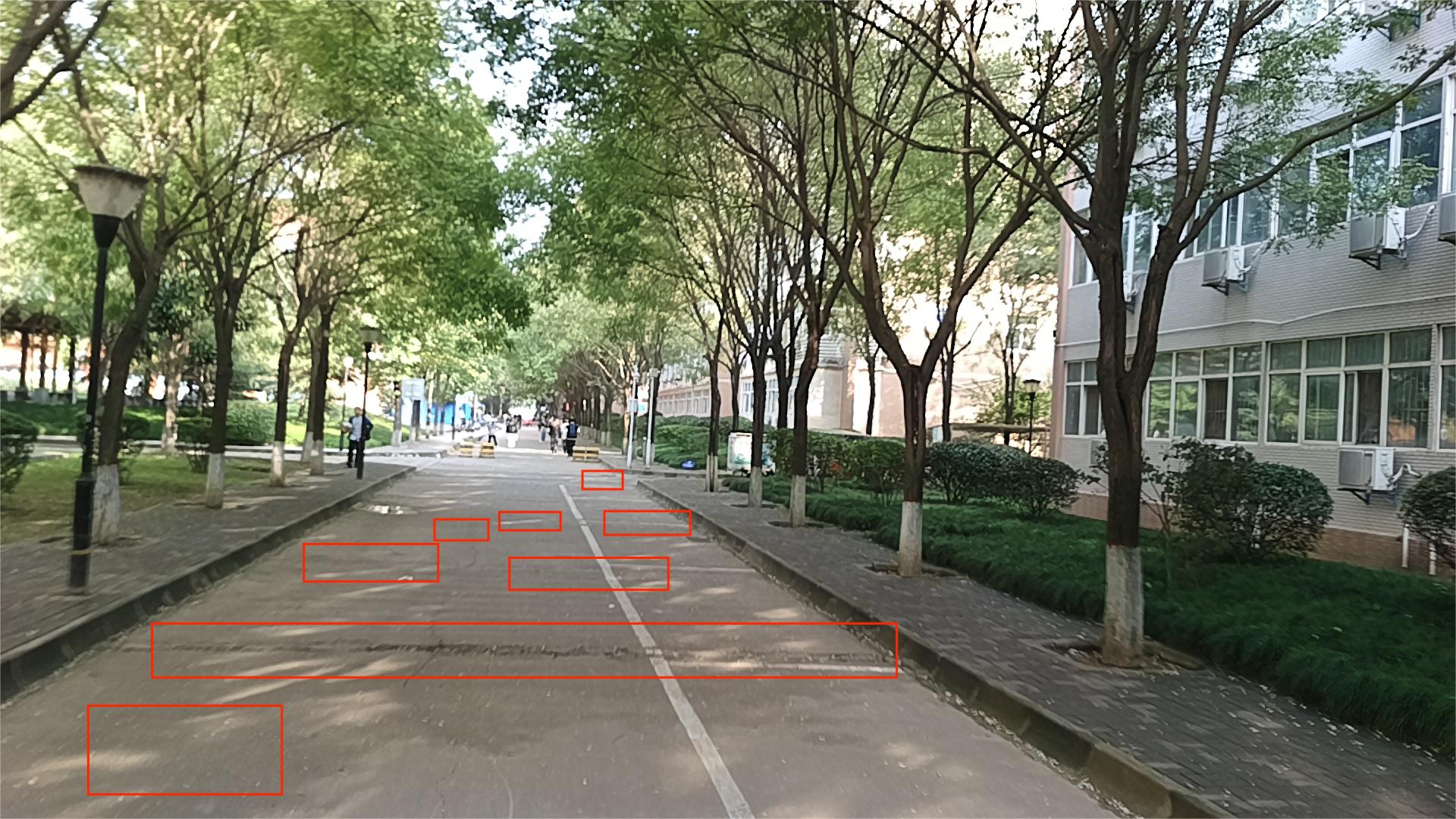}\\
    \hspace*{-0.025\textwidth}Multiple speed bump scenarios \hspace*{0.25\textwidth} Bumpy road sections \\
    \caption{The figure represents our self-collected dataset of speed bump and bumpy road sections, with the speed bumps and bumpy areas highlighted in red boxes.}
    \label{4}
   \end{figure*}  
\subsection{ The System Overview}

Feature points are prone to be lost in frame-to-frame matching due to complexity, while optical flow methods are less likely to lose track in frame-to-frame matching because they rely solely on photometric changes, but the accuracy of optical flow methods is often lower. Therefore, the organic combination of the two makes it possible to achieve high-precision monocular camera SLAM in bumpy scenarios.

ORB-SLAM3AB starts by processing monocular pinhole images, converting them to grayscale, and applying Gaussian denoising. The ORB algorithm extracts feature points, which are then used for optical flow-based frame-to-frame matching. To enhance robustness, we dynamically monitor the number of feature points matched between frames; if the match count is too low, we slightly increase the number of points selected by the optical flow method to ensure frame-to-frame matching can be performed even during bumps. When the match count is sufficient, we reduce the number of points selected by the optical flow method to increase the accuracy of SLAM. 

It should be noted that in ORB-SLAM3, only increasing the number of feature points rather than the optical flow points can lead to unsuccessful frame-to-frame matching or significantly degrade the accuracy of SLAM due to an excessive increase in the number of points. Moreover, there are currently few monocular SLAM systems of this type available for comparative reference.

\begin{itemize}
    \item \textbf{Image Preprocessing}
     
     The input keyframe image (imageKF) and the current frame image (imageF) undergo a color channel check followed by grayscale conversion. If the input images are in RGB format, a conversion function is employed to transform them into grayscale. Subsequently, Gaussian filtering is applied to the grayscale images to reduce noise and enhance the quality of the feature extraction.

    \item \textbf{Feature Point Extraction}
    
     Key points and descriptors are extracted from the grayscale images. The parameters of the feature point extractor, such as the number of detected points, scale factor, and the number of pyramid levels, are adjusted dynamically according to the detection strategy.
     
    \item \textbf{Dynamic Optical Flow Feature Point Detection and Matching}
     
   The detection of feature points using the optical flow method is based on the ORB operator. Initially, the number of optical flow points is set to half the number of feature points. If it is observed that the number of matched feature points in the current frame is insufficient during the subsequent matching process, the number of optical flow feature points is dynamically doubled. The specific values of these parameters should be adjusted according to the actual conditions, especially in scenarios with significant vibrations, where the number of selected feature points may need to be further increased.

   \begin{figure}[htbp]
  \vspace{-10pt}
   \centering
    \subfigure{\includegraphics[width=2cm,height=4cm]{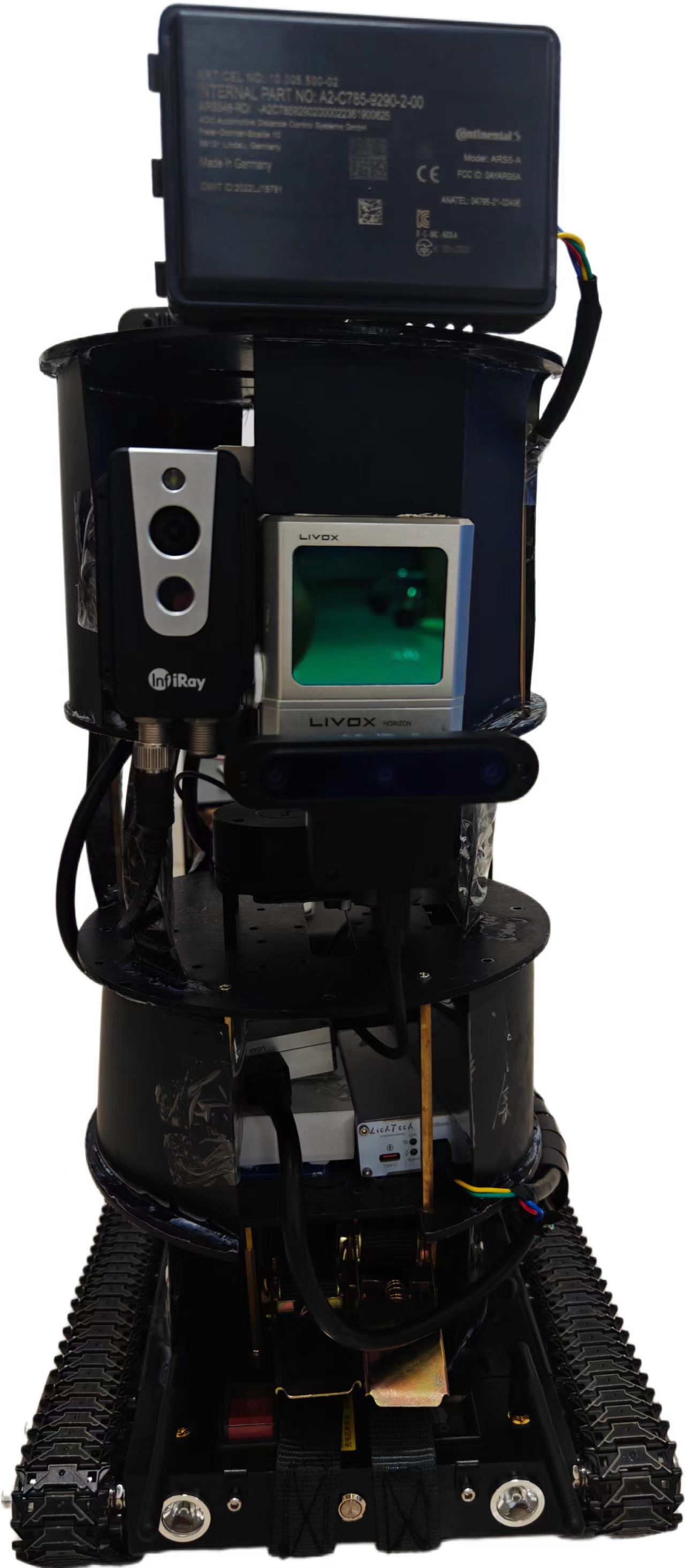}}   \subfigure{\includegraphics[width=6cm,height=4cm]{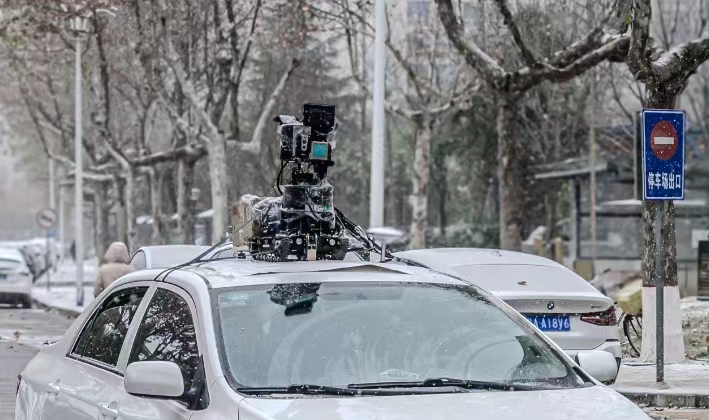}} 
    \subfigure{\includegraphics[width=8.1cm,height=5.9cm]{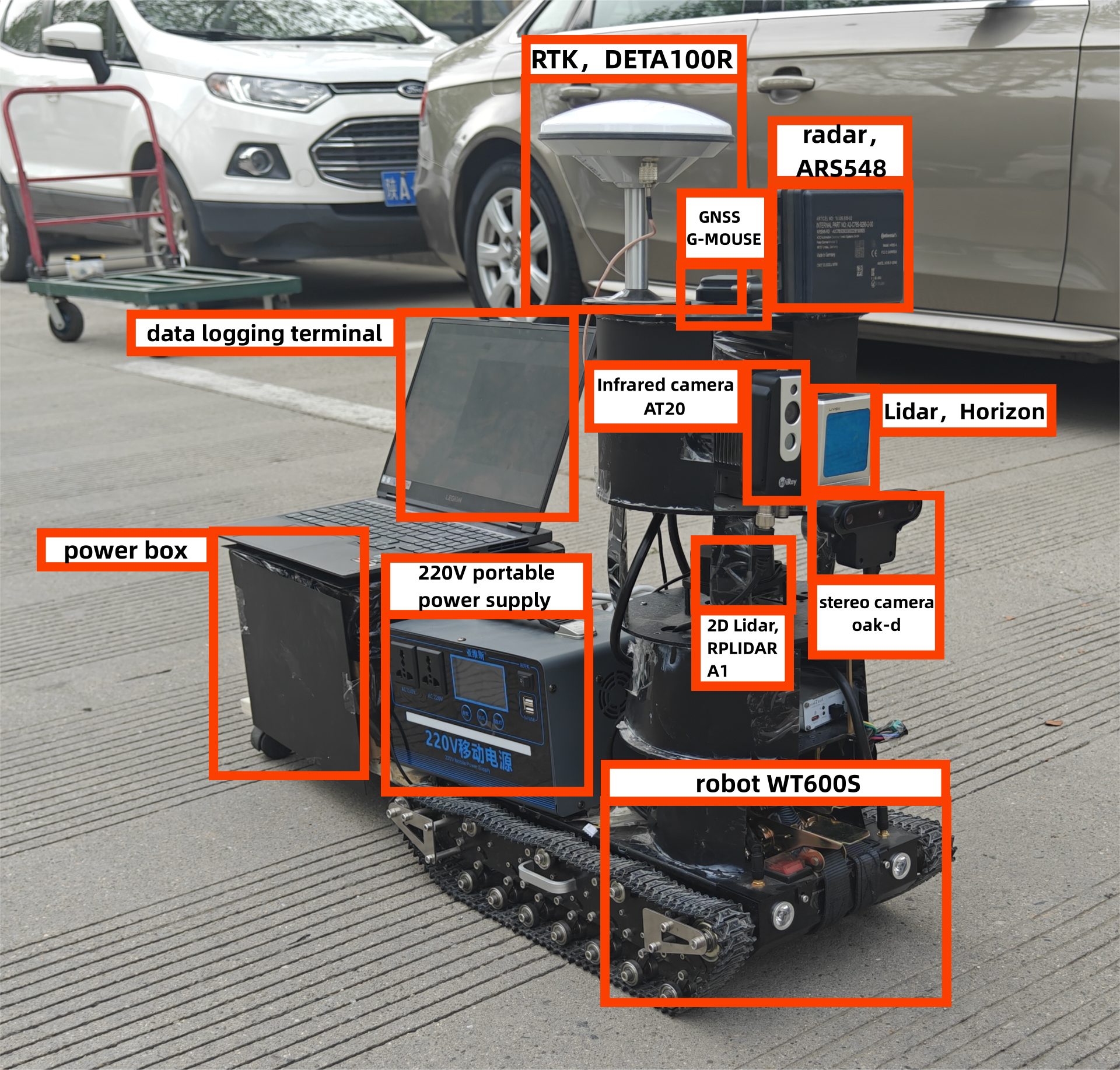}} 
  \caption{The diagram shows the actual ground robot and the sensor-equipped vehicle, with each component’s function clearly labeled.}
    \label{fig:data_distribution1}
    \vspace{-0.2in}
\end{figure}
 
    \item \textbf{Rotation Consistency Check}
    
     A rotation consistency check is performed on the matched feature point pairs. The rotation angles of the feature points are statistically analyzed using the Histogram of Oriented Gradients (HOG) method, and potential erroneous matches are removed from the final results.

\end{itemize}

\begin{figure*}[htbp]
  \vspace{-10pt}
   \centering
    \includegraphics[width=8cm,height=5cm]{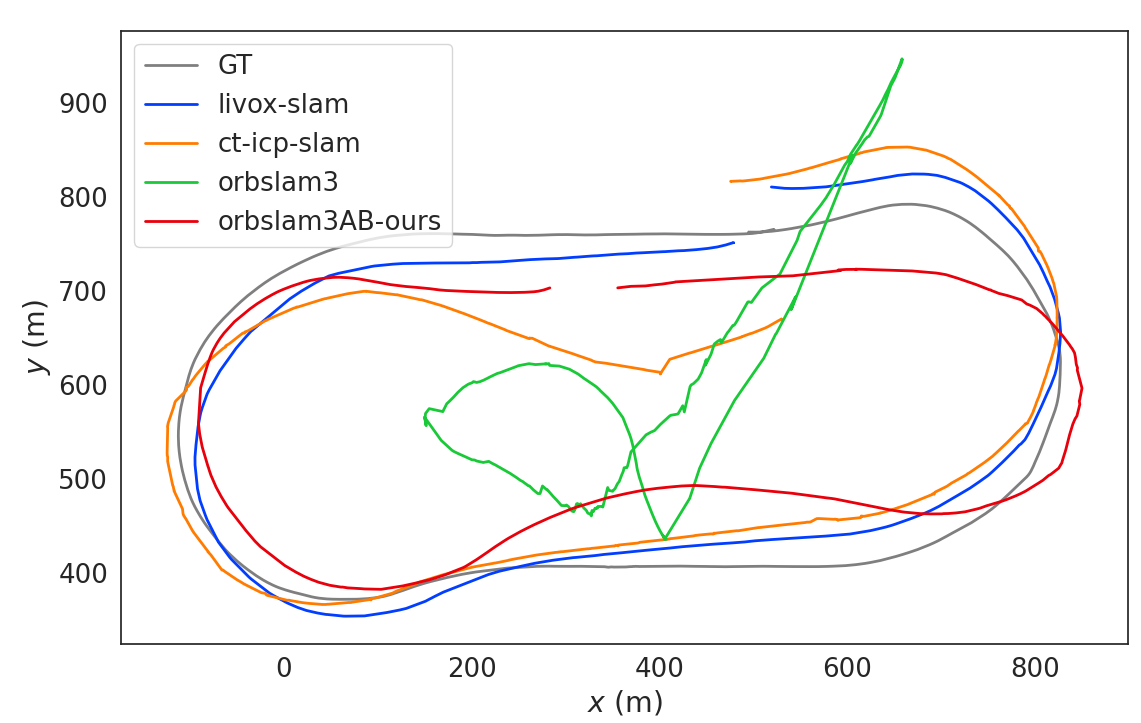} \includegraphics[width=8cm,height=5cm]{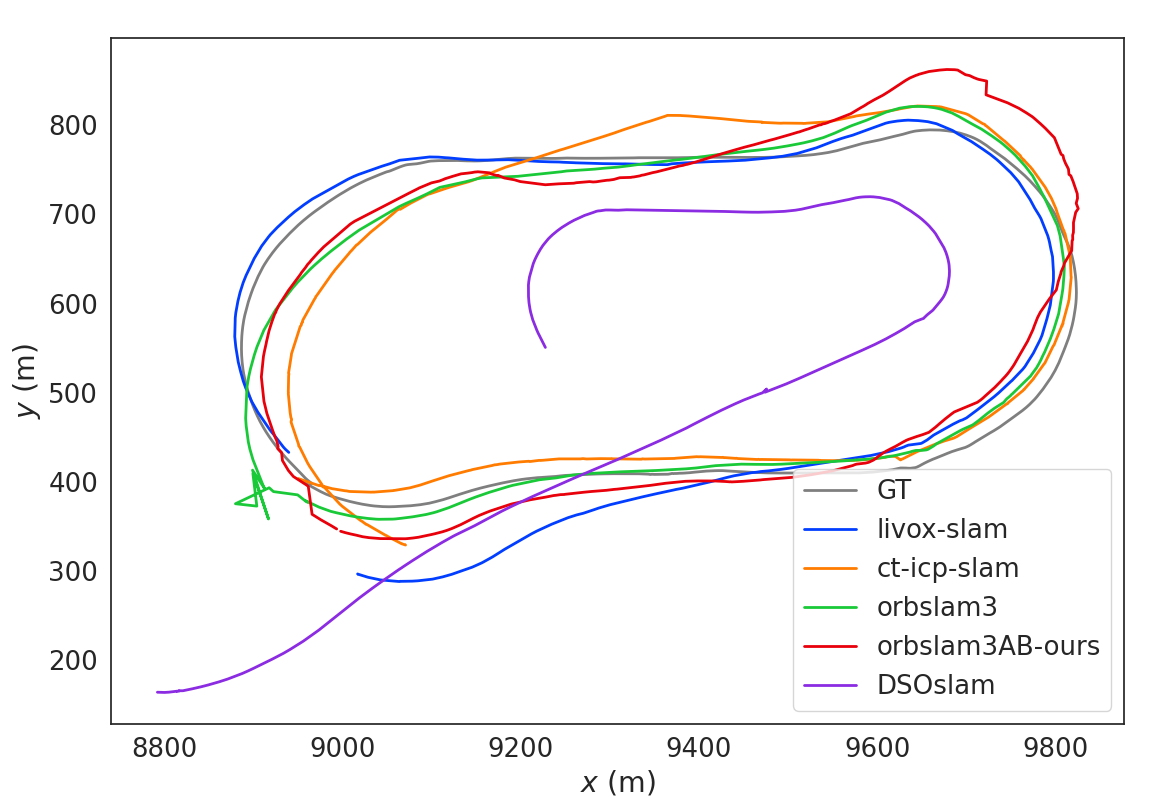}\\
    \hspace{0em}high speed bumps snowy day  \hspace{9em} high speed bumps snowy night \\
    \includegraphics[width=8cm,height=5cm]{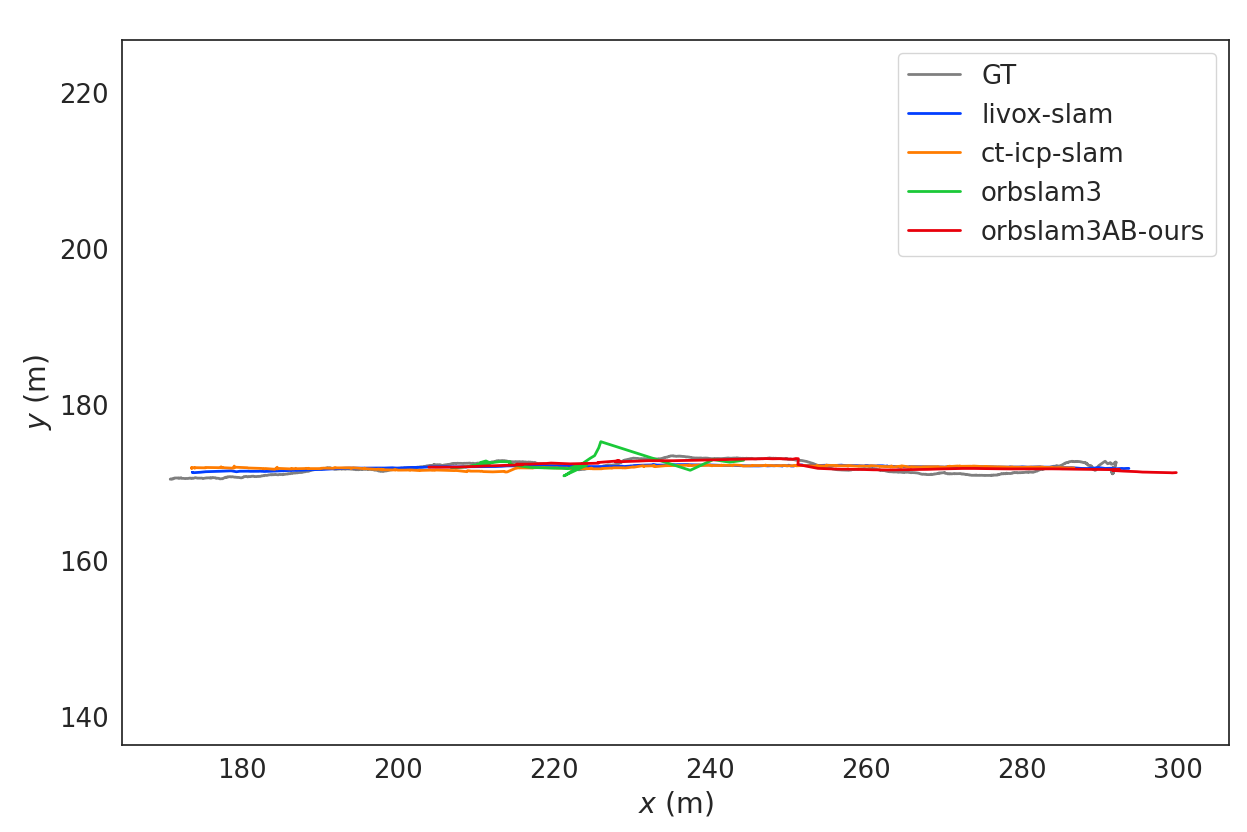}     \includegraphics[width=8cm,height=5cm]{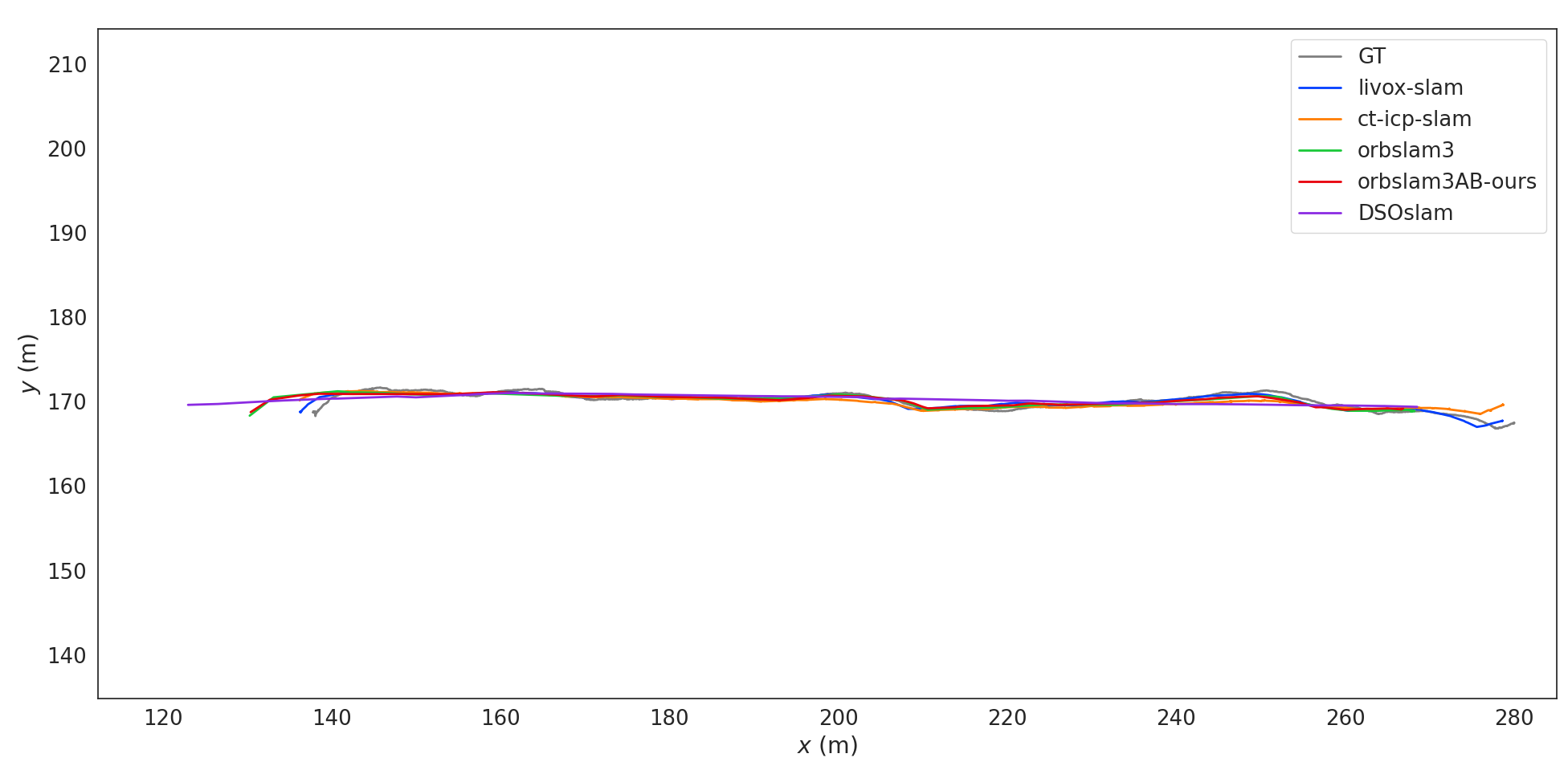}  \\
    \hspace{0em}low speed bumpy sunny day \hspace{9em} low speed bumpy sunny night\\
  \caption{We tested the collected data under various weather, lighting, and road conditions using three laser SLAM algorithms and two visual SLAM algorithms. In high-speed snow conditions with speed bumps, the ORB-SLAM3 algorithm exhibited severe trajectory errors due to intense vibrations. In high-speed snowy night conditions with speed bumps, the DSO-SLAM trajectory was also far from ideal due to the lack of loop closure detection, among other reasons. Our proposed ORB-SLAM3AB algorithm notably outperformed the original ORB-SLAM3 and DSO-SLAM in terms of map accuracy on speed bump surfaces.}
    \label{2}
    \vspace{0.2in}
\end{figure*}

\section{Data Collection}
Due to the rarity of multi-sensor datasets in bumpy scenarios, we designed and assembled a ground robot and integrated it into a vehicle using a detachable mounting system. The robot's structure consists of five layers: the top three layers are equipped with various sensors, while the bottom two layers house interfaces, wiring, mobile power supplies, and computing units. As shown in Figure 3.

To achieve a comprehensive and diverse dataset, we employed both low-speed and high-speed data collection techniques. Our routes included a variety of scenarios, such as bumpy roads and flat roads with multiple speed bumps, and covered various weather conditions, including sunny, rainy, and snowy days, as well as different lighting conditions, including daytime and nighttime. This careful design ensures that our dataset captures a broad spectrum of speed, road, weather, and lighting conditions.

For low-speed data collection, we deployed ground robots to gather data from the same locations under different weather and lighting conditions. This approach allowed us to investigate the effects of specific conditions on sensor performance, with a particular focus on challenging bumpy road segments where loop closure detection was problematic due to their short distances.

In our high-speed data collection phase, we mounted the robot on the roof of a vehicle, maintaining consistent positions while varying weather and lighting conditions. This setup enabled us to collect high-speed data on roads with numerous speed bumps and to record data while navigating bumpy surfaces at high speeds.

We gathered approximately 1500 GB of image and point cloud data. This paper only utilized the bumpy scenarios from the dataset.

\section{ Evaluation }

We applied various SLAM algorithms to the collected data across different scenarios. These included high-speed conditions with multiple speed bumps in sunny, rainy, and snowy weather, as well as low-speed conditions both during the day and night, with and without speed bumps. We evaluated laser-based SLAM and visual SLAM under varying weather conditions, and also assessed visual SLAM performance under different lighting and road conditions. The results were compared against ground truth, providing trajectory maps, Absolute Trajectory Error (ATE)\cite{zhang2018tutorial,shamwell2018vision}, and Relative Pose Error (RPE) \cite{b30,angelidis2014prediction}. Detailed information about these test scenarios is presented in Tables 1 and Figure 4.
\begin{table*}[htbp]
\small\sf\centering
\caption{Comparing ATE and RPE across snowy and bumpy conditions, our ORB-SLAM3AB consistently surpasses both laser-based and visual SLAM methods. "x" marks instances where the algorithm couldn't complete the trajectory.}
\begin{tabular}{lcccccccccc}
\toprule
Sequence & \multicolumn{2}{c}{livox-slam}& \multicolumn{2}{c}{CT-ICP} &\multicolumn{2}{c}{DSO}& \multicolumn{2}{c}{ORB-SLAM3} & \multicolumn{2}{c}{ORB-SLAM3AB(ours)}\\
 & ATE& RPE& ATE& RPE & ATE&RPE & ATE&RPE & ATE&RPE \\
\texttt{low-speed-bumpy-}& && && && && &\\
 \texttt{sunny-day}& 1.845& 0.951& 1.294& 0.568 & x& x& x& x & \textbf{0.228} 
&\textbf{0.099}
\\
 \texttt{low-speed-bumpy-}& & & & & & & & & &\\
\texttt{sunny-night}& 1.901&0.960& 1.852&0.686 & 0.110&0.121& 0.039&0.041 & \textbf{0.038}&\textbf{0.040}
\\
\texttt{high-speed-bumps}& && && && && &\\
 \texttt{snowy-day}& 38.503& 10.920& 44.199& 8.444 & 181.459& 8.044& 4.856& 2.035 & \textbf{4.652} 
&\textbf{0.884}
\\
 \texttt{high-speed-bumps}& & & & & & & & & &\\

 \texttt{snowy-night}& 33.675& 12.819& 53.614& 7.300
 & x&x& 7.643&0.700 & \textbf{1.791} &\textbf{0.283}\\\bottomrule
\end{tabular}\\[10pt]
\end{table*}
Livox-SLAM, CT-ICP represent some of the most advanced laser-based SLAM algorithms available today. These algorithms deliver exceptional accuracy for large-scale SLAM tasks. Our analysis of ATE and RPE demonstrates that laser-based SLAM maintains robust performance over short, uneven terrain. Conversely, in long-distance scenarios with loop closures, the drift errors associated with LiDAR-based SLAM are substantially higher than those observed with visual SLAM methods.

ORB-SLAM3 is a well-established visual SLAM algorithm, known for its rapid response and accurate feature extraction using the ORB operator. However, its robustness significantly decreases under conditions of severe bumps or sharp turns. In high-speed snowy scenarios involving multiple speed bumps, excessive camera shake may lead to feature tracking failures, significant trajectory deviations, and anomalies in loop closure algorithms, resulting in strange-looking rendered images. Similarly, on continuous bumpy low-speed sunny days, the performance of visual SLAM is notably affected.

DSO is a monocular visual SLAM system known for its ability to overcome tracking scenarios with very little texture. This SLAM is more robust in mapping than ORB-SLAM, especially in areas with few features. However, it still exhibits low robustness or low accuracy in scenarios with speed bumps or continuous bumps. In high-speed snowy conditions, due to speed bumps encountered during turns and other reasons, its accuracy is significantly reduced. In high-speed snowy night scenarios, it even loses keyframe information directly, causing the algorithm to be unable to continue working. Similarly, during low-speed continuous bumpy days, the intense shaking of the robot makes feature point selection difficult, leading to an inability to complete the SLAM task.

We employed the improved ORB-SLAM3AB algorithm to test various scenarios, including low-speed bumpy roads during sunny days and nights, as well as high-speed roads with multiple speed bumps on snowy days. From the plotted trajectories, it can be seen that the improved algorithm's tracks more closely match the ground truth. We calculated the corresponding ATE (Absolute Trajectory Error) and RPE (Relative Pose Error). Our algorithm effectively addresses the issue of poor inter-frame matching during bumps, and compared to laser-based SLAM and pure visual SLAM, the new algorithm exhibits lower ATE and RPE values. The experimental results demonstrate that our enhanced algorithm shows greater robustness and improved accuracy on bumpy roads.

\section{Reflection}

Bumpy conditions can significantly degrade the accuracy of both visual SLAM and laser SLAM. Our algorithm leverages the precision of feature point methods and the robustness of optical flow methods to improve ORB-SLAM3 under pure visual conditions, ensuring that it can still complete mapping tasks with relatively high accuracy even in severe bump scenarios. However, introducing optical flow methods might reduce the precision of SLAM on smooth roads. A potential focus for future research could be how to increase robustness in bumpy scenarios without compromising, or even improving, the precision of SLAM on smooth roads. Additionally, investigating whether incorporating LiDAR data during severe bumps through sensor fusion can effectively enhance the robustness of the algorithm is also a direction for future study.

\section{Conclusion}
In this paper, we identified key issues with current visual SLAM systems, particularly the loss of frame-to-frame tracking under strong vibrations and rapid motion. Using ORB-SLAM3 as the framework, we leveraged the robustness of optical flow to increase the detection of flow points during heavy bumps, combining this with the accuracy of feature point methods, which significantly improved the performance of monocular visual SLAM. Due to the scarcity of public datasets that include LiDAR and camera data for speed bumps, continuous bumps, and day and night conditions, we conducted experiments on our self-collected dataset. The experiments demonstrated that our improved algorithm enhances the robustness and accuracy of pure visual SLAM in scenarios with speed bumps or continuous bumps. However, the algorithm still struggles with extremely rapid camera shake and vibrations. Future research will focus on further optimizing the system to address these challenges.

\bibliographystyle{IEEEtran}

\bibliography{IEEEabrv,my.bib}

\end{document}